\def\year{2021}\relax
\documentclass[letterpaper]{article} 
\usepackage{aaai21}  
\usepackage{times}  
\usepackage{helvet} 
\usepackage{courier}  
\usepackage[hyphens]{url}  
\usepackage{graphicx} 
\usepackage{natbib}  
\usepackage{caption}
\usepackage{amsmath}
\usepackage{amssymb}
\usepackage{booktabs}
\usepackage[switch]{lineno}  %

\title{Gaining Insight into SARS-CoV-2 Infection and COVID-19 Severity Using Self-supervised Edge Features and Graph Neural Networks}
\author{Arijit Sehanobish,\thanks{Equal Contribution}\textsuperscript{\rm 1} Neal Ravindra,\footnotemark[1]\textsuperscript{\rm 1} David van Dijk\textsuperscript{\rm 1} \\} 
\affiliations {
\textsuperscript{\rm 1} Internal Medicine (Cardiology) and Computer Science, Yale University \\ \{arijit.sehanobish, neal.ravindra, david.vandijk\}@yale.edu
}
\begin{document}

\maketitle

\begin{abstract}

A molecular and cellular understanding of how SARS-CoV-2 variably infects and causes severe COVID-19 remains a bottleneck in developing interventions to end the pandemic. We sought to use deep learning to study the biology of SARS-CoV-2 infection and COVID-19 severity by identifying transcriptomic patterns and cell types associated with SARS-CoV-2 infection and COVID-19 severity. To do this, we developed a new approach to generating self-supervised edge features. We propose a model that builds on Graph Attention Networks (GAT), creates edge features using self-supervised learning, and ingests these edge features via a Set Transformer. This model achieves significant improvements in predicting the disease state of individual cells, given their transcriptome. We apply our model to single-cell RNA sequencing datasets of SARS-CoV-2 infected lung organoids and bronchoalveolar lavage fluid samples of patients with COVID-19, achieving state-of-the-art performance on both datasets with our model. We then borrow from the field of explainable AI (XAI) to identify the features (genes) and cell types that discriminate bystander vs. infected cells across time and moderate vs. severe COVID-19 disease. To the best of our knowledge, this represents the first application of deep learning to identifying the molecular and cellular determinants of SARS-CoV-2 infection and COVID-19 severity using single-cell omics data. 

  

\end{abstract}

\section{Introduction}

To address the devastating impact of the Coronavirus Disease of 2019 (COVID-19), caused by infection of SARS-CoV-2, and the gap in our understanding of the molecular mechanisms of severe disease and variable susceptibility to infection, we developed a deep learning framework around two single-cell transcriptomic datasets that allowed us to generate hypotheses related to these biological knowledge gaps~\cite{yan_interpretable_covidmort,zhong_immunology_2020}. We rely on single-cell transcriptomic data because single-cell datasets contain rich, molecular and cellular information across a variety of cell types and conditions. We work with two publicly available single-cell datasets: one in which upper airway bronchial epithelium (airway of the lung) organoids were infected with SARS-CoV-2 over a time-course and one dataset of bronchoalveolar lavage fluid samples from patients with varying degrees of COVID-19 severity~\cite{hbec, liao_single-cell_2020}.  Applying machine learning to these datasets allows us to identify the molecular and cellular patterns associated with susceptibility to SARS-CoV-2 infection or severe COVID-19, highlight potential biomarkers, and suggest therapeutic targets. 

Single-cell RNA sequencing (scRNA-seq) is a technology that counts the number of mRNA transcripts for each gene within a single cell~\cite{zheng_scrnaseq_2017, scrnaseq_medicine, review_scrnaseq}. Different tissue samples or cell culture experiments can be assayed with scRNA-seq technology, allowing one to collect information spanning a variety of disease states or perturbations, with thousands of cells' gene expression measured in each experiment. Since transcript counts are correlated with gene expression, scRNA-seq yields large datasets comprising many thousands of cells' gene expression~\cite{zheng_scrnaseq_2017}. However, identifying the genes that determine an individual cell's pathophysiological trajectory or response to viral insult remains a challenge as single-cell data is noisy, sparse, and high-dimensional \cite{compchallenge, rnacluster}. As such, we require cutting-edge deep learning methods to learn how to discriminate cells' controlled experimental perturbation given their transcriptome. Here, we build on previous work that uses graph neural networks (GNNs) to predict an individual cell's disease-state label~\cite{ms_paper}. To reduce bias and improve performance, we developed a novel DL architecture, which, to the best of our knowledge, achieves the highest, single-cell resolved prediction of disease state. Using these models, we then identify the genes and cells important for these predictions. 

GNNs have been widely used and developed for predictive tasks such as node classification and link prediction~\cite{wu_comprehensive_2020}. GNNs learn from discrete relational structure in data but the use of similarity metrics to construct graphs from feature matrices can expand the scope of GNN applications to domains where graph structured data is not readily available~\cite{franceschi_learning_2019,tenenbaum_global_2000}. GNNs typically use message passing, or recursive neighborhood aggregation, to construct a new feature vector for a particular node after aggregating information from its neighbor's feature vectors~\cite{xu_representation_2018,GCN}. 
However recent work~\cite{seshadhri2020impossibility} has shown that these low dimensional node representations may fail to capture important structural details of a graph. Recently, edge features have been incorporated into GNNs to harness information describing different aspects of the relationships between nodes~\cite{gong2018exploiting, gao2018edge2vec, vinyal_message_passing, simonovsky2017dynamic, hu2019strategies} and potentially enrich these low dimensional node embeddings. However, there are very few frameworks for creating \emph{de novo} edge feature vectors that significantly improve the performance of GNNs.
In this article, we propose a self-supervised learning framework to create new edge features that can be used to improve GNN performance in downstream node classification tasks. We hope that our framework provides useful insights into the genes and cell types that might be important determinants of COVID-19 severity and SARS-CoV-2 infection, which can guide further study.

\section{Related work}

There is a large body of research on Graph Neural Networks. A significant amount of work has been focused on graph embedding techniques, representation learning, various predictive analyses using node features and in understanding the representational power of GNNs. There has been recent interest in using edge features to improve the performance of Graph Neural Networks~\cite{gong2018exploiting, chen2019utilizing,Abu_El_Haija_2017, vinyal_message_passing, simonovsky2017dynamic}. However, there are few frameworks to create \emph{de novo} edge features for graphs that do not inherently contain different edge attributes.  

Deep learning, particularly GNNs, have been used in biomedical research to predict medications and diagnoses from electronic health records data~\cite{GRAM}, protein-protein and drug-protein interactions from biological networks, and molecular activity from chemical properties~\cite{nguyen_graphdta_2019, chan_drugdiscoveryAI,Harikumar837807,GAT}. Machine learning has been applied to single-cell data for other tasks, including data de-noising, batch correction, data imputation, unsupervised clustering, and cell-type prediction~\cite{rnacluster,denoising,imputation,cellpred,saucie}. However, fewer works attempt to classify the experimental label associated with each cell or to predict pathophysiological state on an individual cell basis. One recent work uses GAT models to predict the disease state of individual cells derived from clinical samples~\cite{ms_paper}. However, their work does not use edge features. They also do not consider multiple disease states or disease severity. Lastly, they do not account for sample-source bias (i.e., batch effects)~\cite{review_scrnaseq}. In this work, we use a graph-structure that balances neighbors across sample sources to reduce batch effects while preserving biological variation~\cite{bbknn_support}.

Finally there has been a lot of interest in the ML community to interpret black box models. Correctly interpreting ML models can lead to new scientific discoveries and shed light on the biases inherent in the data collection process. 
One of the most common and popular ways to interpret machine learning models is via Shapley values~\cite{lundberg2017unified} and it's various generalizations~\cite{Michalak_2013}. However Shapley values require the independence of features, which is generally hard to guarantee in biological datasets. 
Gradient-based interpretability methods are widely used in computer vision~\cite{ig,DeepLift} and recently, GNNExplainer~\cite{ying2019gnnexplainer} was proposed as a general interpretability method for predictions of any GNN-based model. GNNExplainer identifies a compact sub-graph structure and a small subset of node features that play an important role in a network's prediction. It is a gradient-based method and the authors formulate it as an optimization task that maximizes the mutual information between a GNN's prediction and the distribution of possible sub-graphs. In this work, in addition to GNNExplainer, we follow the approach of~\cite{ms_paper,attn_state_space_schaar} in using attention mechanisms for interpretability.

To the best of our knowledge, this is the first attempt to apply a GNN architecture to gain insight into multiple pathophysiological states at single-cell resolution, merging time-points, severity, and disease-state prediction into a multi-label node classification task.

    

\begin{figure*}[ht]
  \centering
  \includegraphics[width=0.9\textwidth]{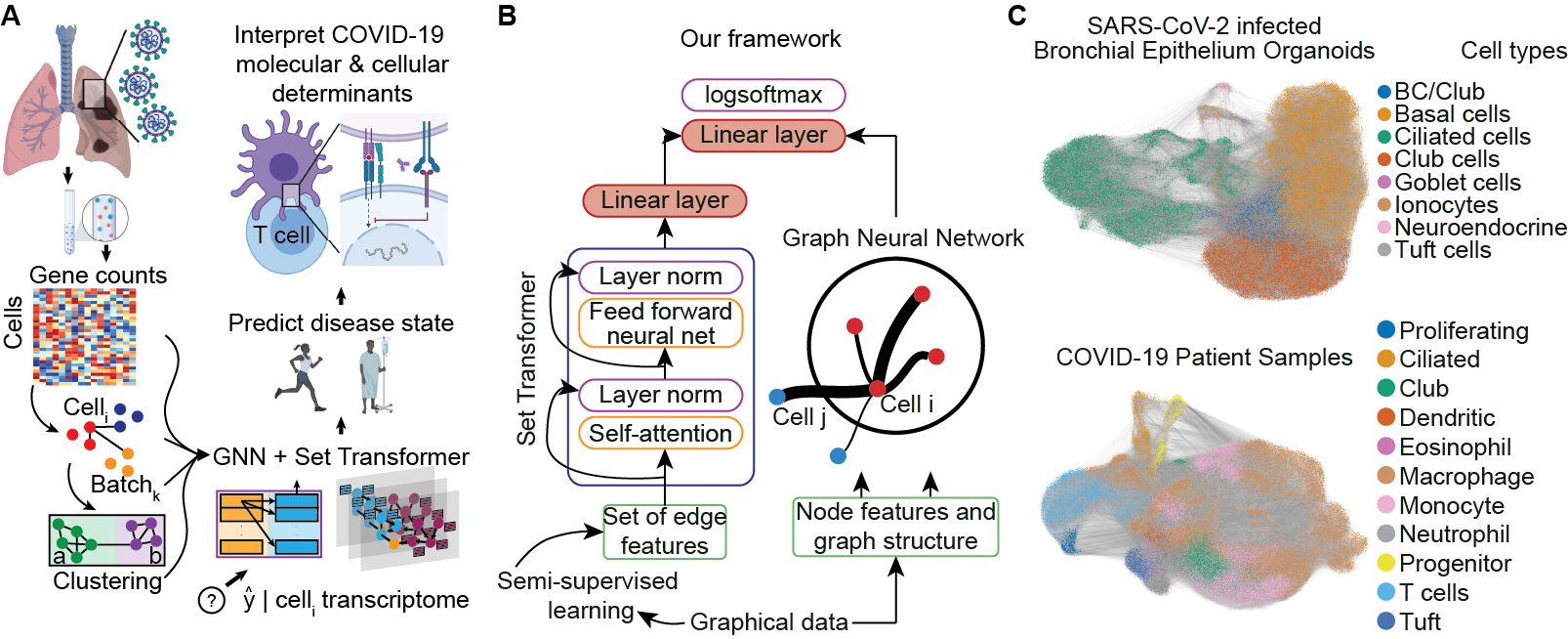}
  \caption[0.95\textwidth]{Our framework and datasets of interest. (\textbf{A}) Overview of our approach with respect to gaining molecular and cellular insights into COVID-19. (\textbf{B}) Our framework and model architecture, integrating edge features with GNNs via a Set Transformer. (\textbf{C}) Graphical data used, showing cell types for each cell and edges in a dimensionality-reduced embedding.}
  \label{fig:model}
\end{figure*}

\section{Models}

In this section we describe our model, which consists of two components: (1) A Set Transformer and (2) Graph Attention Network (GAT) layers.

\subsection{Set Transformer}

We use a Set Transformer as in~\cite{lee2018set}. The Set Transformer is permutation invariant so it is an ideal architecture to encode sets. The building block of our Set Transformer is the multi-head attention, as in \cite{Attention}. Given $n$ query vectors $Q$ of dimension $d_q$, a key-value pair matrix $K \in \mathbb{R}^{n_v \times d_q}$ and a value matrix $V \in \mathbb{R}^{n_v \times d_v}$ and, assuming for simplicity that $d_q = d_v=d$, the attention mechanism is a function given by the following formula: 
\begin{equation}\label{eqn:settrans_attn_coeff}
    \text{att}(Q,K,V) := \text{softmax}(\frac{QK^{T}}{\sqrt{d}})V
\end{equation}
This multihead attention is computed by first projecting $Q,K,V$ onto $h$ different $d^{h}_q, d^{h}_q, d^{h}_v$ dimensional vectors where, for simplicity, we take $d^{h}_q = d^{h}_v = \frac{d}{h}$ such that,
\begin{equation}
\text{Multihead}(Q, K, V) := \text{concat}(O_1, \cdots , O_h)W^O
\end{equation}
where 
\begin{equation}
O_j = \text{att}(QW_{j}^{Q}, KW_{j}^{K},VW_{j}^{V}) \end{equation}, and $W_{j}^{Q}, W_{j}^{K},W_{j}^{V}$ are projection operators of dimensions $\mathbb{R}^{d_q \times d^{h}_q}, \mathbb{R}^{d_q \times d^{h}_q}$ and $\mathbb{R}^{d_v \times d^{h}_v}$, respectively, and $W^O$ is a linear operator of dimension $d \times d$. Now, given a set $S$, the Set Transformer Block (STB) is given the following formula: 
\begin{equation}
    STB(S):=  \text{LayerNorm}(X + rFF(X))
\end{equation}
where 
\begin{equation}
X = \text{LayerNorm}(S + \text{Multihead}(S, S, S))
\end{equation}
and rFF is a row-wise feedforward layer and LayerNorm is layer normalization~\cite{ba2016layer}.

Given a set of elements with input dimension $d_{in}$, the Set Transformer outputs a set of the same size with output dimension $d_{out}$. Since we will be dealing with sets of variable lengths, instead of outputting sets, we aggregate the output vectors to produce a single dense vector of dimension $d_{out}$. In particular, if for some set $S$ of $n$ elements, $\{w_1, \cdots ,  w_n \}$ is the output of the Set Transformer for the set $S$, we use learnable weights $\lambda_j$ to combine the vectors via the following equation :
\begin{equation}
    w : = \sum_{j=1}^{n}\lambda_{j}w_{j}
\end{equation}

\subsection{Graph Attention Network}
We use the popular Graph Attention Network (GAT) as the backbone to learn node representations and also for creating edge features based on our auxiliary tasks. We follow the exposition in~\cite{GAT}. The input to a GAT layer are the node features, $\mathbf{h} = \{ h_1, h_2, . . . , h_N \}$, where $h_i \in \mathbb{R}^F $, $N$ is the number of nodes, and $F$ is the number of features in each node. The layer produces a new set of node features (of possibly different dimension $F'$) as its output, $\mathbf{h'} = \{h'_1,h'_2,....h'_N \}$ where $h'_i \in \mathbb{R}^{F'}$.  The heart of this layer is multi-head self-attention like in~\cite{Attention, GAT}. Self-attention is computed on the nodes,
 \begin{equation}\label{eqn:feed_forward}
     a^{l} : \mathbb{R}^{F'} \times \mathbb{R}^{F'} \rightarrow \mathbb{R}
 \end{equation}
where $a^l$ is a feedforward network. Using self-attention, we can obtain attention coefficients,
\begin{equation} \label{eqn:unnormalized_attn}
e^{l}_{ij} = a^{l}(\mathbb{W}^{l}h_{i}, \mathbb{W}^{l}h_j )
\end{equation}
where $\mathbb{W}^{l}$ is a linear transformation and also called the weight matrix for the head $l$.
We then normalize these attention coefficients. 
\begin{equation}\label{eqn:att_coeff}
\alpha^{l}_{ij} = \text{softmax}_{j} (e^{l}_{ij} ) = \frac{\text{exp}(e^{l}_{ij} )} {\sum_ {k \in \mathcal{N}_{i}} \text{exp}(e^{l}_{ik})}
\end{equation}
where $\mathcal{N}_i$ is a $1$-neighborhood of the node $i$. The normalized attention coefficients are then used to compute a linear combination of features, serving as the final output features for each corresponding node (after applying a nonlinearity, $\sigma$):
\begin{equation} \label{eqn:new_feat}
h^{l}_{i} = \sigma \bigg( \sum_{j \in \mathcal{N}_i} \alpha^{l}_{ij}\mathbb{W}^{l}h_{j} \bigg).
\end{equation}
We concatenate the features of these heads to produce a new node feature, $h'_{i} := \left\vert\right\vert h^{l}_{i}$.

However, for the final prediction layer, we average the representations over heads and apply a logistic sigmoid non-linearity. Thus the equation for the final layer is: 
\begin{equation}
    h'_{i} = \sigma \bigg(\frac{1}{K}\sum_{l=1}^{K} \sum_{j \in \mathcal{N}_i} \alpha^{l}_{ij}\mathbb{W}^{l}h_{j} \bigg).
\end{equation}
where $K$ is the number of heads. 

Based on auxiliary tasks, our new edge features $\Lambda_{ij}$ for the edge $e_{ij}$ are created by concatenating the $\alpha^{l}_{ij}$ across all heads, i.e. 
\begin{equation}\label{eqn:new_edge_feat}
\Lambda_{ij} := \lvert\rvert_{l=1}^{K} \alpha^{l}_{ij}
\end{equation}

\subsection{Our model}

In this subsection we will describe our model that combines edge features with node features for our main node classification task. We use two GAT layers to encode the node representations. In the case of the GAT layers, we concatenate the representations obtained by different heads resulting in a $64$-dimensional node feature vector. For each node $i$, we construct a set $S_{i}: = \{ \Lambda_{ij}: j \in N_{i} \}$, where $\Lambda_{ij}$ is the vector representing the edge features of the edge $e_{ij}$ connecting the nodes $i$ and $j$. We then encode this set, $S_{i}$, which we call the edge feature set attached to the node $i$ via our Set Transformer. We use $2$ heads and $1$ block of the Set Transformer, outputting a $8$-dimensional vector. This $8$-dimensional vector is concatenated with the $64$-dimensional node representation from the GAT layer. We call this new representation an enhanced node feature vector. This enhanced node feature vector is then passed through a dense layer with a logistic sigmoid non-linearity for classification. More details about the model and the training hyperparameters can be found in the Appendix. We finally note that instead of GAT layers, we can also use any message passing GNN layers in our main node classification task.

\section{Datasets Used}
We validate our model on the following scRNA-seq datasets: 
\begin{itemize}
    \item 4 human bronchial epithelial cell cultures or ``organoids" that were inoculated with SARS-CoV-2 and co-cultured for 1, 2, and 3 days post-infection~\cite{hbec}.
    \item Bronchoalveolar lavage fluid samples from 12 patients enrolled in a study at Shenzen Third People's Hospital in Guangdong Province, China of whom 3 were healthy controls, 3 had a mild or moderate form of COVID-19 and 6 had a severe or critical COVID-19 illness~\cite{liao_single-cell_2020}.
\end{itemize}

\begin{table}[h]
\centering
\caption{Dataset description showing train/val/test splits.}
\label{tab:tab1_datasets}
\resizebox{\columnwidth}{!}{%
\begin{tabular}{@{}cccc@{}}
\toprule
Datasets & \begin{tabular}[c]{@{}c@{}}SARS-CoV-2 \\ infected organoids\end{tabular} & COVID-19 patients \\ \midrule
\# Nodes         & 54353/11646/11648     & 63486/13604/13605          \\
\# Node features & 24714                 & 25626                                   \\
\# Edges         & 1041226/230429/228630 & 2746280/703217/707529  \\
\# Edge features & 18                    & 18    \\
\# Classes & 7                    & 3           \\ \bottomrule
\end{tabular}%
}
\end{table} 
\textbf{Data Preprocessing :} All single-cell samples were processed with the standard scRNA-seq pre-processing pipeline using Scanpy~\cite{Scanpy,satija2015}. To create graphs from a cell by gene counts feature matrix, we used a batch-balanced, weighted kNN graph~\cite{polanski_bbknn_2019}. BB-kNN constructs a kNN graph that identifies the top $k$ nearest neighbors in each ``batch", effectively aligning batches to remove sample-source bias while preserving biological variability~\cite{bbknn_support}. We used annoy's method of approximate neighbor finding by calculating Euclidean distances between nodes in $50$-dimensional PCA space. The PCA space is obtained by dimensionality-reduction (via principal component analysis) of the normalized and square-root transformed cell by gene counts matrix. Per ``batch" we find $k=3$ nearest neighbors, with edge weights given by the distance between corresponding nodes. An example BB-kNN graph is schematized in Figure \ref{fig:model}A. 

\textbf{Single-cell label creation :} For the COVID-19 patient dataset, all cells from each patient sample were given labels in accordance with that patient’s COVID-19 severity (healthy, moderate, or severe). For the organoid dataset, cells with more than 10 transcripts aligned to the SARS-CoV-2 genome were considered to be infected. Cells in the 1, 2, and 3 days-post-infection (dpi) samples that were not infected are bystander cells. Mock is a control sample and can’t be bystander or infected. The 3 timepoints were concatenated to the infection label per cell to yield 7 labels across the dataset (Mock, 1dpi-infected, 1dpi-bystander, and so on). Given the proximity of bronchoalveolar lavage fluid cells to the primary site of viral insult, we make the assumption that the transcriptomes of cells from a COVID-19 patient indicate response to disease. Thus, all cells from one patient have the same label. Similarly, we assume that all cells in an organoid culture inoculated with SARS-CoV-2 exhibit transcriptomic signatures associated with being an infected or bystander cell, distinct from mock-infected or control sample cells. 

\textbf{Model performance :} To generate train/test/val sets, we pooled all cells from a single dataset, then randomly assigned 70/15/15\% of cells to train/test/val. We created a separate batch-balanced kNN graph for each split. To mini-batch the graphs, we used the Pytorch Geometric implementation of the ClusterData algorithm~\cite{clusterloader}. The validation set was used to select the model and the trained classifier was evaluated on the unseen test set. We evaluate the model based on node label accuracy. The negative log-likelihood loss is computed with respect to the the ground truth label of the nodes, derived from sample metadata (as described above).


\begin{table}[h]
\centering
\caption{Experimental tasks}
\label{tab:tab2_tasks}

\resizebox{\columnwidth}{!}{%
\begin{tabular}{l c c c } 
\toprule
Task & \begin{tabular}[c]{@{}c@{}}SARS-CoV-2\\ infected organoids\end{tabular} & COVID-19 patients  &  \\ \midrule
Louvain cluster ID     & Cell type               & Cell type                                 &  \\
Batch or node metadata & Culture sample ID       & Patient ID                   &  \\
Inductive prediction   & Timepoint and infection & No, Mild, or Severe Disease                      &  \\ \bottomrule
\end{tabular}
}
\end{table}


\section{Creating new edge features}
In this section we describe our method to create new edge features. 
\subsection{Creating new edge features via auxilliary tasks}
\textbf{Predicting Louvain clusters via GAT :} We cluster our datasets using Louvain clustering~\cite{louvain}, and annotate these clusters as ``cell types", as commonly done in single-cell analysis~\cite{rnacluster}. More information about these tasks, e.g., the number of clusters, can be found in the Appendix. Then, we use a $2$-layer GAT with $8$ attention heads in each layer to predict the cell type label. We extract the edge attention coefficients from the first layer of our trained model to use as edge features in our main node classification task. Thus we get an $8$-dimensional edge feature vector by equation~\ref{eqn:new_edge_feat}. \\
\textbf{Predicting other metadata associated to our graphs :} All of our biological datasets have a batch or sample ID associated to it, i.e. some metadata that keeps track of the origin of the cell. We use the same method as before to create another $8$-dimensional edge feature vector. More details and results about the auxiliary tasks can be found in the Appendix. 

\subsection{Creating dataset agnostic features}
In this subsection we quickly describe some other methods to create new edge features. \\
\textbf{Forman-Ricci curvature :}
We now use the internal geometry of the graph to create our next edge feature. We use the discrete version of the Ricci curvature as introduced by Forman~\cite{Forman} and discussed in~\cite{Samal_2018}:
\begin{align*}\label{forman-ricci}
Ric_{F}(e) &:= \omega(e) \bigg(\frac{\omega(v_1)}{\omega(e)} + \frac{\omega(v_2)}{\omega(e)}  \\
&- \sum_{e_{v_1} \sim v_1, e_{v_2} \sim v_2}\bigg[\frac{\omega(v_1)}{\sqrt{\omega(e)\omega(e_{v_1})}} + \frac{\omega(v_2)}{\sqrt{\omega(e)\omega(e_{v_2})}}\bigg]\bigg)
\end{align*}
where $e$ is an edge connecting the nodes $v_1$ and $v_2$, $\omega(e)$ is the weight of the edge $e$, $\omega(v_i)$ is the weight of the node, which we take to be $1$ for simplicity, and $e_{v_i} \sim v_i$ is the set of all edges connected to $v_i$ and \textit{excluding} $e$. This is an intrinsic invariant that captures the local geometry of the graph and relates to the global property of the graph via a Gauss-Bonnet style result~\cite{watanabe2017combinatorial}. We found that our graphs are hyperbolic and most of the edges are negatively curved. As a future work, we would like to employ the methodologies introduced in~\cite{Albert_2014} to understand how the hyperbolicity affects higher order connectivities and the biological implications of such connectivities. We further hope that their methods would shed light on the most relevant paths between source and target nodes and to identify the most important nodes that govern these pathways. \\
\textbf{Edge features via node2vec :}
We use a popular embedding method called node$2$vec~\cite{grover2016node2vec} to embed the nodes in a $64$ dimensional space. We then calculate the dot product between these node embeddings as a measure of similarity. However to be consistent with our other methods, we only compute the dot product between the nodes which share an edge. node2vec embeddings preserve the local community structure of a graph, which we expect should provide information to enhance discriminability between nodes, as previously suggested~\cite{khosla_comparative_2019}.

Finally we concatenate all the created vectors into an $18$ dimensional edge feature vector which we use in our main node classification task.

\section{Experiments}
Our main task is node classification in an inductive setting, as shown in Table~\ref{tab:tab2_tasks}. We compare our model and framework against popular GNN architectures like ClusterGCN~\cite{clusterloader,GCN} and GAT~\cite{GAT} as well as different set encoding frameworks like DeepSet~\cite{zaheer2017deep} and Set2Set~\cite{vinyals2015order}. We also compare our model against GNN models that incorporate edge features like Graph Isomorphism Network, as modified in~\cite{hu2019strategies}, and a Dynamic Edge Conditioned Convolution Network (ECC)~\cite{simonovsky2017dynamic, vinyal_message_passing}. All the results shown are from the test set and our model's performance is reported in table~\ref{tab:tab3_main}. Our model outperforms the baseline models by a significant margin and also outperforms the other state-of-the-art networks and frameworks. Our processed data and code can be found at \url{https://github.com/nealgravindra/self-supervsed_edge_feats}.

\begin{table}[h]
\centering
\caption{Accuracy and $95\%$ confidence intervals for $n=6$ trials except for models marked with$^{1}$, where $n=3$.}
\label{tab:tab3_main}
\resizebox{.9\columnwidth}{!}{%
\begin{tabular}{@{}cccc@{}}
\toprule
Models &
  \begin{tabular}[c]{@{}c@{}}SARS-CoV-2\\ infected organoids\end{tabular} &
  \begin{tabular}[c]{@{}c@{}}COVID-19\\ patients\end{tabular}  \\ \midrule
ClusterGCN                       & 65.43 (65.21-65.65)          & 89.26 (89.06-89.47)              \\
ClusterGCN + DeepSet & 79.75 (78.75-80.75) & 87.2 (87.02-87.38) \\
ClusterGCN + Set2Set & 71.65 (69.89-73.42) & 88.34 (87.89-88.79) \\
ClusterGCN + Set Transformer & 81.61 (79.34-83.87) & 92.84 (91.95-93.74)  \\
GAT                        & 73.10 (70.93-75.27)          & 92.25 (91.27-93.24)  \\
GAT + DeepSet & 79.45 (77.98-80.92) & 75.99 (74.8-77.68) \\
GAT + Set2Set & 82.95 (81.75-84.15) & 92.87 (92.62-93.12) \\
GAT + Set Transformer (Ours) & \textbf{89.8 (88.89-91.71)} & \textbf{95.12 (94.02-96.22)}\\ 
GIN + EdgeConv$^1$ & 63.36 (62.53-64.19 & 89.56 (88.54-90.58) \\
EdgeConditionedConvolution$^1$ & 46.15 (34.72-57.59) & 88.63 (86.07-91.20)  \\
\bottomrule
\end{tabular}%
}
\end{table}

\section{Ablation Studies}

In this section we sought to understand how our edge features affect model performance. 
\begin{table}[h]
\centering
\caption{Ablation studies showing accuracy. Row names corresponding to the first column indicate which edge feature has been used with our model (GAT + Set Transformer).}
\label{tab:my-table}
\resizebox{\columnwidth}{!}{%
\begin{tabular}{ccc}
\toprule
Edge Feature & SARS-CoV-2 infected organoids & COVID-19 patients  \\ \midrule
  Cluster label & .7137 & .9211 \\ 
  Batch label & .8381 & .9264  \\
  node2vec & .6976 & .9111 \\
  Curvature & .7205 & .9215 \\
  Cluster + batch label & .8449 & .9689 \\
node2vec + curvature & .6929 & .9168 \\
 Cluster + batch label + node2vec & .8443 & .9602 \\
 Cluster + batch label + curvature & .8438 & .9605 \\ \bottomrule
\end{tabular}%
 }
\end{table}
A more detailed list of ablation studies can be found in the Appendix. From table~\ref{tab:my-table}, we can see that edge features derived from the cell types and batch ID improve the model performance.

\begin{figure*}[ht!]
  \centering
  \includegraphics[width=.9\linewidth]{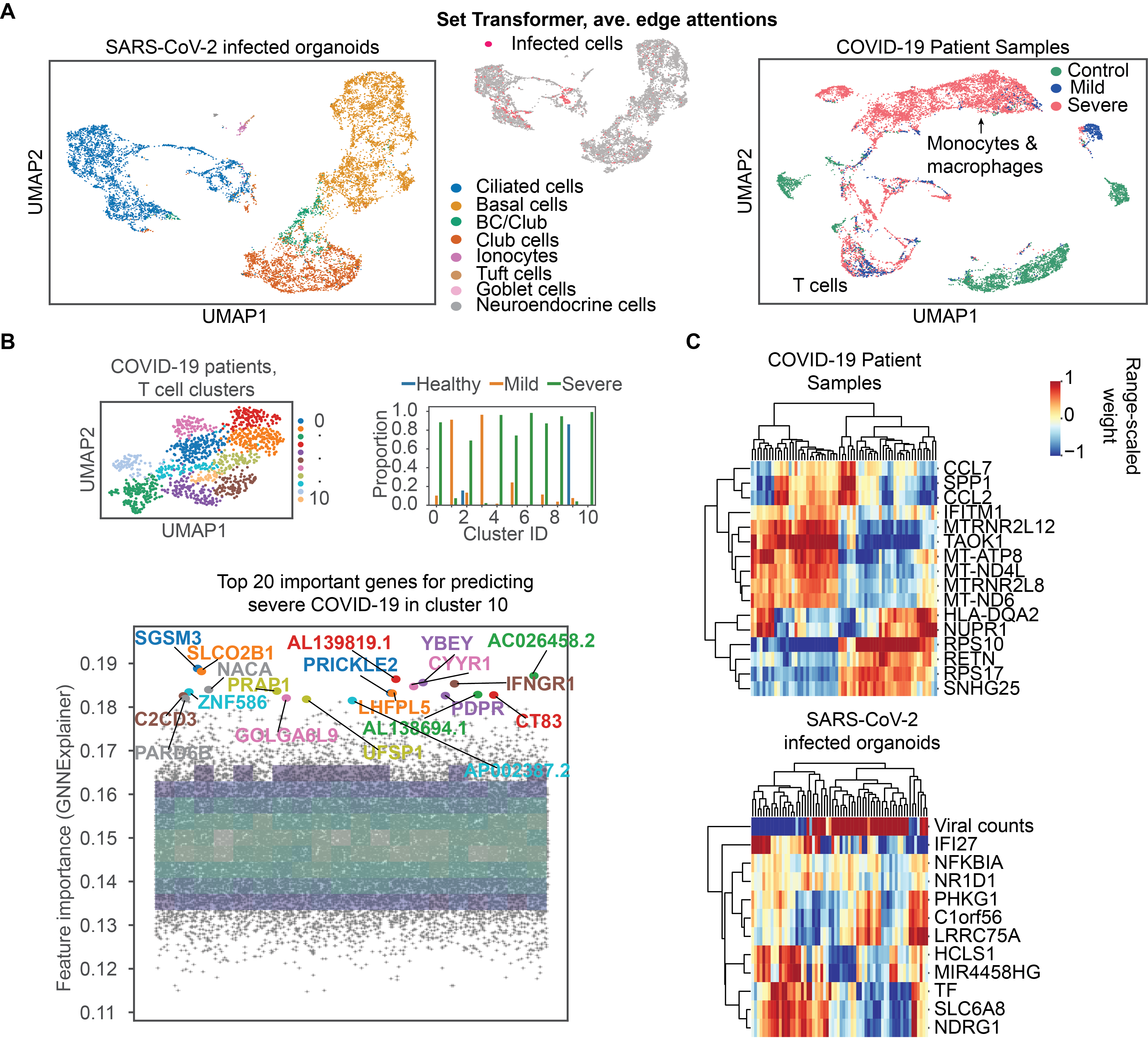}
  \caption[0.95\textwidth]{Model interpretability and the genes and cells important to COVID-19 severity and susceptibility to SARS-CoV-2 infection. (\textbf{A}) Embeddings learned from graphs constructed by averaging the edge attention coefficients across Set Transformer output dimensions, showing cell type or condition per cell. (\textbf{B}) Top 20 important genes for predicting a prototypical cell in a severe COVID-19 patient T cell cluster using GNNExplainer. If the proportion of points in a 20x20 grid is over 0.1, then point density is shown as color in a heatmap. (\textbf{C}) Top 5 important gene features for each GAT head, colored by learned weights.}
  \label{fig:interpretability}
\end{figure*}

\section{Discussion \& Interpretability}

We sought to gain insight into biological mechanism by extracting how our model learned to distinguish the different transcriptional signatures of SARS-CoV-2 infection dynamics and COVID-19 disease severity. We show the various aspects of model interpretability that we can glean from our model in Figure \ref{fig:interpretability}.

First, we extract the learned, edge attention coefficients from our Set Transformer
and average these across attention heads to yield 1-dimensional edge weights. We use those edge weights to construct a new adjacency matrix. Then, with a cosine distance metric and this new adjacency matrix, we learn a new embedding of the cells (Figure \ref{fig:interpretability}A) using the UMAP algorithm~\cite{umap}. In addition, to evaluate the importance of different types of edge features, we plot the average weights of the query matrix in the Set Transformer (see Appendix). Using the attention coefficients for manifold learning shows better separation of cells by cell type and label than typically used for embedding high-dimensional single-cell data (where the input for manifold learning is a cell by gene counts matrix), possibly because our model accounts for cell type variability via their edge feature representation. These embeddings may be useful in identifying unique, phenotypic subsets of cells. For example, in the new cell embedding of SARS-CoV-2 infected organoids, ciliated cells overlap with infected cells in a distinct and dense cluster. Indeed, it is thought that SARS-CoV-2 preferentially infects ciliated cells, which suggests that this type of model interpretability may have some utility~\cite{hbec}. In the new cell embedding of COVID-19 patient samples, the T cell population is mixed with cells derived from patients with mild and severe COVID-19, while a cluster predominantly comprised of cells derived from patients with severe COVID-19 is made up of macrophages and monocytes (\ref{fig:interpretability}A, right). T cells and monocytes derived from macrophages are important in regulating the immune response and are targets for a number of therapies currently under development~\cite{checkpointinhibitors, liao_single-cell_2020, iwasaki2020}. Furthermore, T cells are regulated by interferon signaling, which is itself a current COVID-19 therapeutic target~\cite{interferon_drug}. Taken together, this suggests that our model may identify cellular subsets worthy of further study to complement existing biomedical research. 

After finding that T cells in severe COVID-19 patients may be hard to distinguish from healthy or mildly afflicted COVID-19 patients, we sub-clustered T cells from the test set and identified a prototypical T cell (cell nearest to cluster 10's centroid in UMAP space) in a cluster unique to COVID-19 patients with severe disease (\ref{fig:interpretability}B). This allowed us to identify the most important features for predicting disease severity in this unique severe COVID-19 patient T cell cluster using a gradient-based approach (GNNExplainer)~\cite{ying2019gnnexplainer}. Expected genes, i.e., genes thought to play a role in immunopathology associated with COVID-19 severity, arose in the top 20 most important genes, such as genes involved in interferon signaling and inflammation (IFNGR1, SLCO2B1). However, some novel genes also arose, for example, related to cardiac remodeling and metabolic regulation (NACA, ZNF586, PDPR, PRICKLE2, C2CD3, SGSM3, PARD6B, AL139819), which may suggest a unique response to SARS-CoV-2 infection or a cardiovascular component of severe COVID-19, the latter of which has been clinically suggested~\cite{iwasaki2020, clinicalCOVIDriskFactors}.

We also extract the learned weights (the matrices $\mathbb{W}^{l}$ for $l=1,\cdots, 8$) from our models' first GAT layer and average them over $8$ heads in order to globally investigate our model's feature saliency and indicate which genes are important in discriminating between transcriptomes of patients with varying degrees of COVID-19 severity and of lung cells with variable susceptibility to SARS-CoV-2 infection. In predicting COVID-19 severity from patient samples, our model gives high weight to genes involved in the innate immune system response to type I interferon (CCL2, CCL7, IFITM1), regulation of signaling (NUPR1, TAOK1, MTRNR2L12), a component of the major histocompatibility complex II (HLA-DQA2), which is important for developing immunity to infection, and a marker of eosinophil cells (RETN), a cell type involved in fighting parasites and a suspected source of immunopathology during COVID-19~\cite{iwasaki2020}. In predicting SARS-CoV-2 infection, our model has saliency for viral transcript counts, which is encouraging, as well as genes that are involved in the inflammatory response and cell death (NFKBIA), as well as signaling (IFI27, HCLS1, NDRG1, NR1D1, TF), which may provide clues as to the dynamic regulatory response of cells in the host's lungs to SARS-CoV-2.

\section{Conclusion} 


Here, we attempt to bring accurate disease state prediction to a molecular and cellular scale so that we can identify the cells and genes that are important for determining susceptibility to SARS-Cov-2 infection and severe COVID-19 illness via model interpretability. We achieved significant improvements in accuracy compared to other popular GNN architectures with our framework. Additionally, relative to vanilla GNNs, we achieve better separation of cells by cell type when visualizing the attention coefficients than possible with GATs alone. We also hypothesize that by computing edge features using the cell type and batch label, we control for these factors of variation in our main classification task and thus obtain potentially more meaningful features associated with COVID-19 than other models.

This suggests that using edge features derived from self-supervised learning can improve performance on disease-state classification from single-cell data. We used our models to gain insights into the cell tropism of SARS-CoV-2 and to elucidate the genes and cell types that are important for predicting COVID-19 severity. 
It is interesting that our model finds that genes involved in type I interferon signaling are important in predicting both COVID-19 severity and susceptibility to SARS-CoV-2 infection. It is suspected that dysregulation of type I interferon signaling may cause immunopathology during SARS-CoV-2 infection, leading to critical illness~\cite{hbec, iwasaki2020}. Further study into the interaction partners and the subtle transcriptional differences between the genes and cells that we identified as important for prediction may provide complementary hypotheses or avenues for therapeutic intervention to mitigate the impacts of COVID-19. However, we are not medical professionals so we do \emph{NOT} claim that interpretation of our model will bear any fruit. Rather, we hope that the approach of seeking state-of-the-art results on predicting disease states at single-cell resolution will enhance the study of biology and medicine and potentially accelerate our understanding of critical disease.

\section{Ethical statement}
There are many caveats to our study. While we achieve good performance with our models, model interpretability in artificial neural networks does not have a strong theoretical basis, and any proposed important features should merely be thought of as putative biological hypotheses. In addition, the cells in our datasets are derived from a relatively small patient population. While we attempt to limit sample-source bias by using a batch-balanced graph, we remain vulnerable to the idiosyncrasies of our samples. Thus, any putative hypotheses should only be considered meaningful after experimental validation. 

\section*{Acknowledgements}
We acknowledge the Yale Center for Research Computing for our use of their High Performance Computing infrastructure. We thank the anonymous reviewers for helpful comments and suggestions. We also thank Mia Madel Alfajaro and Craig B. Wilen for generating the SARS-CoV-2 infected organoids dataset and sharing the data with us.

\bibliography{ref}
\vspace{1cm}
\appendix

\section{Data pre-processing}

\subsection{Feature matrix preparation}

Prior to graph creation, all samples were processed with the standard single-cell RNA-sequencing pre-processing recipe using Scanpy~\cite{Scanpy,satija2015}. For the SARS-CoV-2 infected organoids and COVID-19 patients datasets, genes expressed in fewer than 3 cells and cells expressing fewer than 200 genes were removed but, to allow for characterization of stress response and cell death, cells expressing a high percentage of mitochondrial genes were not removed. For all single-cell datasets, transcript or "gene" counts per cell were normalized by library size and square-root transformed. 

\subsection{Graph creation}

To create graphs from a cell by gene counts feature matrix, we used a batch-balanced kNN graph~\cite{polanski_bbknn_2019}. BB-kNN constructs a kNN graph that identifies the top $k$ nearest neighbors in each "batch", effectively aligning batches to remove bias in cell source while preserving biological variability~\cite{bbknn_support}. We used annoy's method of approximate neighbor finding with a Euclidean distance metric in $50$-dimensional PCA space. Per "batch" we find $k=3$ top nearest neighbors. An example BB-kNN graph is schematized in main text, Figure 1A.  
\begin{figure}[h]
  \centering
  \includegraphics[width=\columnwidth]{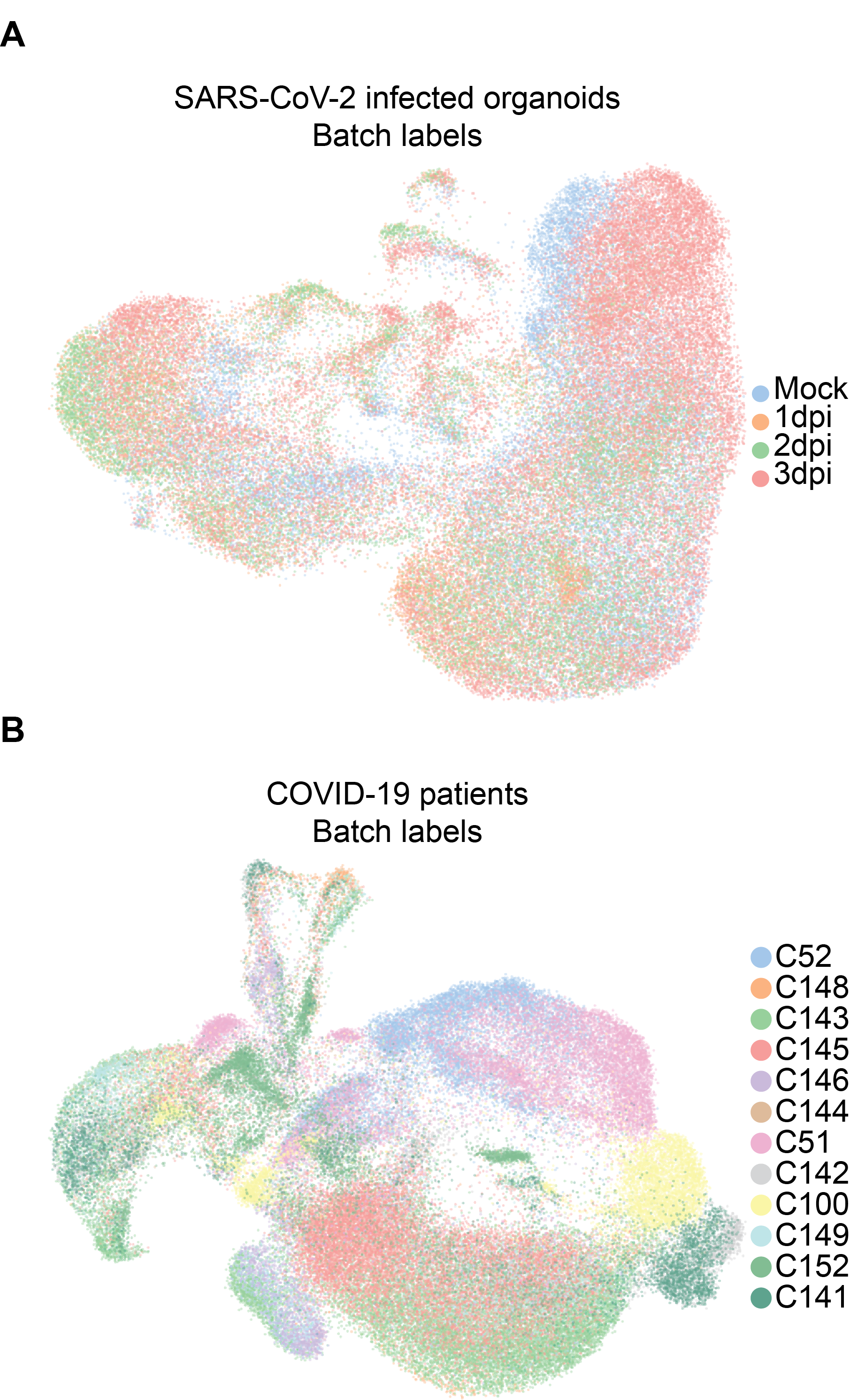}
  \caption[0.95\textwidth]{UMAP embeddings of individual cells colored by labels for auxiliary tasks. (\textbf{A}) Batch labels for SARS-Cov-2 infected organoids dataset. (\textbf{B}) Batch labels for COVID-19 patients, for patient IDs described in~\cite{liao_single-cell_2020}.}
  \label{fig:cell_type}
\end{figure}


\section{Hyperparameters and Training details}
\begin{table}[h]
\centering
\caption{Default hyperparameters used in the experiments}
\resizebox{\columnwidth}{!}
{
\begin{tabular}{ccc}
\hline
 & Graph Attention Network & Graph Convolution Network \\
\hline
Number of layers & $2$ & $2$ \\

Hidden\_size & $8$ & $256$ \\

Attention Heads & $8$ & N/A \\

Optimizer & Adagrad & Adagrad\\


weight\_decay & $.0005$ & $.0005$ \\

Batch size & $256$ & $256$\\

Dropout & $.5$ & $.4$ \\

Slope in LeakyRelu & $.2$ & $.2$ \\

Training Epochs & $1000$ & $1000$ \\

Early stopping &  $100$ & $100$\\
\hline
\end{tabular}

\label{tab:hyperparameters}
}
\end{table}

For auxiliary tasks and for training our models, we break our graph into $5000$ subgraphs using the ClusterData function in PyTorch Geometric library and then minibatched the subgraphs using the ClusterData function. These algorithms are originally introduced in~\cite{clusterloader}. We used a single block of Set Transformer with input dimension $18$, output dimension $8$ and $2$ heads. The rest of the hyperparamaters of GAT and GCN can be found in table~\ref{tab:hyperparameters}. 

For our auxiliary tasks and for baseline experiments we used an 8GB Nvidia RTX2080 GPU and for our main tasks we used an Intel E5-2660 v3 CPU with 121GB RAM.

\section{Auxiliary task}
In this section we describe our auxiliary tasks. Table~\ref{tab:aux_label} gives details about the number of labels for the auxiliary tasks. We first predict the cell types as given by the Louvain clustering~\cite{louvain}. In the main text, we used~\cite{marker_db} to obtain cell type markers and annotate the Louvain cluster labels as "cell types" explicitly. 

Next we predict the batch ID of each node, i.e. which patient or from where the cell is obtained. Table~\ref{tab:aux_task} shows our results for these auxiliary tasks. 
In single-cell RNA-sequencing, variability between
batches can explain more of the transcriptomic variability than variability in the biological process of interest; these "batch effects" can complicate model inference~\cite{rnacluster}. Our novel use of BB-kNN graphs for the tasks described in this paper limits this "batch effect" bias. 


\begin{table}[h!]
\centering
\caption{Number of labels for auxiliary tasks}
\label{tab:aux_label}

\begin{tabular}{l c c c c} 
\toprule
Task & \begin{tabular}[c]{@{}c@{}}SARS-CoV-2\\ infected organoids\end{tabular} & COVID-19 patients & \\ \midrule
Cell type  &      8         &        10           &    \\
Batch  &   4    &   12   \\
\bottomrule
\end{tabular}
\end{table}

\begin{table}[h!]
\centering
\caption{Results on auxiliary tasks}
\label{tab:aux_task}

\resizebox{\columnwidth}{!}{%
\begin{tabular}{l c c c } 
\toprule
Prediction & \begin{tabular}[c]{@{}c@{}}SARS-CoV-2\\ infected organoids\end{tabular} & COVID-19 patients &   \\ \midrule
Cell type  &      93.84         &        82.03           &       \\
Batch  &   76.16    &   64.08     \\
\bottomrule
\end{tabular}
}
\end{table}

\section{Ablation Studies}
In this section we show detailed ablation studies to understand how the combination of edge features that we created affects our model performance. 
\begin{table}[h]
\centering
\caption{\tiny Ablation studies showing accuracy. Row names corresponding to the first column indicate which edge feature has been used with our model (GAT + Set Transformer).}
\label{tab:full_ablation_tab}
\resizebox{.9\columnwidth}{!}{%
\begin{tabular}{ccc}
\toprule
Edge Feature & SARS-CoV-2 infected organoids & COVID-19 patients  \\ \midrule
  Cluster label & .7137 & .9211 \\ 
  Batch label & .8381 & .9264  \\
  node2vec & .6976 & .9111 \\
  Curvature & .7205 & .9215 \\
  Cluster + batch label & .8449 & .9689 \\
node2vec + curvature & .6929 & .9168 \\
 Cluster + batch label + node2vec & .8443 & .9602 \\
 Cluster + batch label + curvature & .8438 & .9605 \\ 
 Cluster + curvature & .7149 & .9199\\
 Cluster + node2vec & .7536 & .9169 \\
 Batch label + curvature & .833 & .944 \\
 Batch label + node2vec & .8569 & .9344 \\
 Cluster + curvature + node2vec & .6991 & .9164 \\
 Batch + curvature + node2vec & .8091 & .9489 \\
 Set Transformer Averaged & .8535 & .9433 \\
 \bottomrule

\end{tabular}%
 }
\end{table}
Finally equation (6) in the main text shows how we combine the output of the Set Transformer into a single dense vector using learnable weights. We sought to understand the importance of these learnable weights by removing the linear layer and averaging the output of the Set Transformer. The final row of table~\ref{tab:full_ablation_tab} shows the results by ablating the linear layer with just a simple average.


\end{document}